\documentclass[letterpaper, 10 pt, conference]{ieeeconf}  

\IEEEoverridecommandlockouts                              
\usepackage{cite}
\usepackage{amsmath,amssymb,amsfonts}
\usepackage{algorithmic}
\usepackage{graphicx}
\usepackage{textcomp}
\usepackage{subfigure}
\usepackage{caption}
\usepackage{subcaption}
\usepackage[nolist]{acronym}
\usepackage{siunitx}
\renewcommand{\SIrange}[3]{\lbrack\num{#1}\ {;}\ \num{#2}\rbrack\,\si{#3}}

\usepackage{hhline}
\usepackage{multirow}
\usepackage{bm}
\usepackage{pifont}
\usepackage{hyperref}

\let\labelindent\relax
\usepackage{enumitem}
\usepackage{array}
\usepackage{flushend}   
\usepackage{wrapfig, lipsum}

\usepackage[usenames,dvipsnames]{xcolor}
\usepackage{tikz}
\usepackage{tkz-tab}
\usetikzlibrary{patterns}

\usepackage{pgfplots}
\usepackage{pgfplotstable}
\pgfplotsset{compat=1.9}
\usepgfplotslibrary{external}

\newcommand{\ie}{i.\,e., }      

\hypersetup{
    pdftitle={Hybrid Machine Learning Model with a Constrained Action Space for
Trajectory Prediction},
    pdfdisplaydoctitle={false},
    pdfauthor={Alexander Fertig, Lakshman Balasubramanian and Michael Botsch},
    pdfsubject={Hybrid Machine Learning Model with a Constrained Action Space for
Trajectory Prediction},
    pdfkeywords={Autonomous Driving, Representation Learning, IV, Deep Learning, Machine Learning, DL, ML, Latent Space, Loss},%
}

\begin{acronym}
    \acro{av}[AV]{Autonomous Vehicle}
	\acroplural{av}[AVs]{Autonomous Vehicles} 
    \acro{bev}[BEV]{bird's-eye-view}
	\acro{ml}[ML]{Machine Learning}
	\acro{dl}[DL]{Deep Learning} 
	\acro{cnn}[CNN]{Convolutional Neural Network} 
	\acro{lstm}[LSTM]{Long Short-Term Memory} 
	\acro{rnn}[RNN]{Recurrent Neural Network} 
    \acro{dkm}[DKM]{Deep Kinematic Model}
    \acro{kf}[KF]{Kalman-Filter}
    \acro{gnn}[GNN]{Graph Neural Network}
	\acroplural{gnn}[GNNs]{Graph Neural Networks} 
    \acro{fde}[FDE]{Final Displacement Error}
    \acro{ade}[ADE]{Average Displacement Error}
    \acro{wmm}[WMM]{Without Motion Model}
    \acro{hms}[HMS]{Hybrid Multi Shot}
    \acro{hss}[HSS]{Hybrid Single Shot}
    \acro{cv}[CV]{Constant Velocity}
    \acro{ctra}[CTRA]{Constant Turn Rate and Acceleration}
    \acro{mse}[MSE]{Mean Squared Error}
    \acro{gft}[GFT]{Graph Fourier Transform}
\end{acronym}

\newcommand\copyrighttext{%
  \footnotesize \textcopyright 2025 IEEE. Personal use of this material is permitted. Permission from IEEE must be obtained for all other uses, in any current or future media, including reprinting/republishing this material for advertising or promotional purposes, creating new collective works, for resale or redistribution to servers or lists, or reuse of any copyrighted component of this work in other works.
  DOI: \href{https://doi.org/10.1109/IV64158.2025.11097759}{10.1109/IV64158.2025.11097759}}
\newcommand\copyrightnotice{%
\begin{tikzpicture}[remember picture,overlay]
\node[anchor=south,yshift=10pt] at (current page.south) {\fbox{\parbox{\dimexpr\textwidth-\fboxsep-\fboxrule\relax}{\copyrighttext}}};
\end{tikzpicture}%
}

\title{\LARGE \bf
Hybrid Machine Learning Model with a Constrained \\ Action Space for Trajectory Prediction
}

\author{Alexander Fertig$^{*1}$, Lakshman Balasubramanian$^{*2}$ and Michael Botsch$^{1,2}$
\thanks{*Equal contribution}
\thanks{$^{1}$Technische Hochschule Ingolstadt, AImotion Bavaria, Esplanade 10, 85049 Ingolstadt, Germany {\tt\small firstname.lastname@thi.de}}%
\thanks{$^{2}$Technische Hochschule Ingolstadt, Research Center CARISSMA, Esplanade 10, 85049 Ingolstadt, Germany {\tt\small extern.lakshman.balasubramanian@thi.de}}%
}

\begin{document}

\bstctlcite{IEEEexample:BSTcontrol}

\maketitle
\copyrightnotice
\thispagestyle{empty}
\pagestyle{empty}

\begin{abstract}
Trajectory prediction is crucial to advance autonomous driving, improving safety, and efficiency. Although end-to-end models based on deep learning have great potential, they often do not consider vehicle dynamic limitations, leading to unrealistic predictions. 
To address this problem, this work introduces a novel hybrid model that combines deep learning with a kinematic motion model.
It is able to predict object attributes such as acceleration and yaw rate and generate trajectories based on them.
A key contribution is the incorporation of expert knowledge into the learning objective of the deep learning model.
This results in the constraint of the available action space, thus enabling the prediction of physically feasible object attributes and trajectories, thereby increasing safety and robustness.
The proposed hybrid model facilitates enhanced interpretability, thereby reinforcing the trustworthiness of deep learning methods and promoting the development of safe planning solutions.
Experiments conducted on the publicly available real-world Argoverse dataset demonstrate realistic driving behaviour, with benchmark comparisons and ablation studies showing promising results.

\end{abstract}

\section{Introduction}
	
Trajectory prediction of traffic participants in traffic scenario is a crucial aspect of autonomous driving.
In recent years, the task of trajectory prediction has been addressed largely using \ac{dl}-based methods like~\cite{Gao2020,Zhao2020, Janjos2022}.  
The success of \ac{dl} methods can be attributed to their ability to combine map information with traffic participants and to model complex interactions between traffic participants. 
Network architectures are chosen depending on the input data representation such as occupancy grids~\cite{Nadarajan2023}, graphs~\cite{Neumeier2022} or polylines and the network is tasked with predicting the future motion of the traffic participants for a fixed prediction time. 
This task is usually trained end-to-end using real-world driving data. 

Such end-to-end trained models are unconstrained trajectory prediction models.
Therefore, it is possible that the predicted trajectories or object states do not meet the physical constraints of vehicle dynamics. 
This issue is addressed mainly in \ac{dkm}~\cite{Cui2020}, StarNet~\cite{Janjos2022} and SafetyNet~\cite{Vitelli2022}. 
All of these works propose hybrid models combining \ac{dl}- and model-based approaches. 
Thus, the \ac{dl} model is tasked with predicting the action space representing object states, i.\,e., the acceleration/deceleration or yaw rate for the traffic participants. 
The motion model uses the action space created by the \ac{dl} model to predict the future trajectories of traffic participants. This architecture can be trained in an end-to-end fashion.

All the hybrid models discussed above make the implicit assumption that optimising for the predicted trajectories will optimise the action space predicted by the \ac{dl} method as well. 
However, the action space predicted by the \ac{dl} model is not bound by the physical constraints of vehicle dynamics and therefore does not provide valid safety guarantees.

To address the aforementioned limitations, this work proposes a hybrid model for trajectory prediction that can be used to include the physical constraints of vehicle dynamics on the action space. 
The hybrid model consists of two main components: a modern \ac{dl} model and a traditional motion model.
Firstly, the \ac{dl} model processes the scenario information and generates an output within an intermediate action space.
In order to limit the action space output of the \ac{dl} model, the learning objective is constrained by incorporating physical constraints based on expert knowledge with respect to the traffic participants.
Secondly, the motion model then uses these intermediate action space outputs to predict the concrete trajectories of the traffic participants within the scenario.
In this work, the task of trajectory prediction includes the prediction of object attributes, also referred to as motion prediction. 
The overall architecture of the proposed hybrid model is trained in an end-to-end fashion. 
The incorporation of physical constraints within the learning objective has been demonstrated to be a valuable component within the proposed architectural framework for the generation of realistic action space outputs and driveable trajectories. The main contributions are as follows.
\begin{itemize}
    \item Development of a hybrid model for trajectory prediction, such that:
    \begin{enumerate}[label=(\alph*)]
        \item \ac{dl} and motion models are combined in a trainable end-to-end manner,
        \item a physical motion model predicts drivable trajectories,
        \item the action space is formed by incorporating physical constraints of vehicle dynamics, which are based on expert knowledge and formulated within the learning objective.
    \end{enumerate}
    \item The importance of considering physical constraints is demonstrated through an evaluation of the proposed hybrid method, conducted through an ablation study and a benchmark comparison utilising the publicly available real-world Argoverse dataset. 
\end{itemize}

\section{Related Work}
\label{Sec:related_works}

The field of trajectory prediction has been an active area of research in recent years. 
Model-based methods, such as~\cite{Treiber2017, Wagner2020, Elter2022} used physical motion models to capture the underlying vehicle dynamics.
Over the years, various \ac{dl}-based architectures~\cite{Gao2020}, \cite{Zhao2020}, \cite{Neumeier2022}, \cite{Neumeier2023b}, \cite{Song2022}, \cite{Neumeier2024} have been proposed, which mainly differ in how traffic scenario information is encoded and how interactions and predictions are modelled. 
Hereby, \ac{dl} methods usually generate trajectories for a pre-determined time in the future.

However, the generated trajectories may lack physical feasibility, as no explicit physical constraints are defined during the training process.
This may result in the generation of vehicle trajectories that are not physically drivable \cite{Cui2020}.

Hybrid models can address the issue of the kinematic feasibility of the predictions and ensure drivable trajectories, two types of hybrid models are suggested in current research. 
The first type is a two-step approach, as proposed in~\cite{PhanMinh2020}. 
In a first step, a set of physically feasible trajectories is generated using a model-based method, based on the current vehicle state. 
In the second step, a neural network is employed to determine the most probable trajectory from this previously generated set of trajectories.
This is a computationally expensive method during inference and training, as the set of physically possible trajectories is generated and evaluated for every scenario. 

The second type of hybrid models comprises an alternative two-step architecture, which combines learning-based models with vehicle motion models.
In the first step, the machine learning model is employed to generate the action space inputs for the motion models such as acceleration, yaw rate, etc. 
In the second step, the motion model utilises the outputs from the machine learning model to generate the trajectories.
Thereby, the physical drivability of the trajectories is ensured by the motion model.
One of the first methods to introduce this concept is the \ac{dkm} \cite{Cui2020}. 
The authors propose to use a \ac{cnn} and \ac{rnn} backbone combined with a physical motion model to predict future trajectories of the vehicle. 
The \ac{cnn} and the \ac{rnn} backbone predict features within the action space, \ie acceleration and yaw rate. 
These predictions are used as input for the motion model to predict the trajectories in the next time step. 
The work of \cite{Girase2021} extends \ac{dkm} with a feasible path-conditioned encoder to keep the vehicle on possible routes. 

The methods proposed in~\cite{Janjos2022} and~\cite{Vitelli2022} are also using a kinematic model to generate trajectories instead of directly predicting the trajectories from the machine learning model. 
Although all of these methods use kinematic models in trajectory prediction, none of the aforementioned works directly incorporates the action space outputs into the learning objective, except~\cite{Vitelli2022}. 
They rely on the implicit assumption that optimising for the predicted trajectories will also produce optimal action space outputs.
Nevertheless, as demonstrated experimentally in this work, the action space outputs may be physically infeasible despite the trajectory outputs aligning with the learning objectives. 
This represents a current limitation of many hybrid model approaches, which could potentially lead to safety constraints.
In order to address this issue, this work proposes a hybrid model in which expert knowledge is incorporated into the learning objective to shape the action space in a manner that aligns with the principles of vehicle dynamics.

\section{Methodology}
\label{Sec:methodology}
\begin{figure*}
    \centering
    \vspace{0.12cm}
    \includegraphics[width=0.825\textwidth]{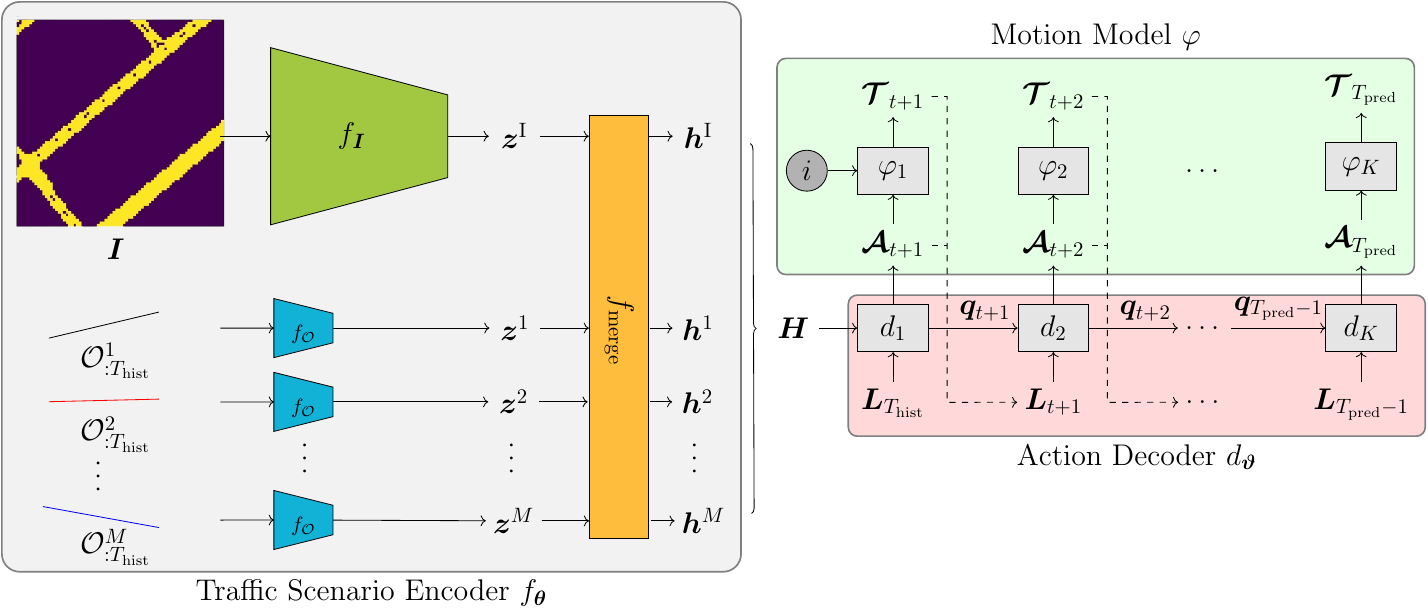}
    \caption{Architecture overview of the proposed hybrid model $g$. The traffic scenario encoder $f_{\bm{\theta}}$ creates a latent representation of the given scenario. The action decoder $d_{\bm{\vartheta}}$ predicts the action space outputs for the traffic participants, and hereby \mbox{$\bm{L}_{t+1} = (\bm{\mathcal{A}}_{t+1}, \bm{\mathcal{T}}_{t+1})$}. The motion model $\varphi$ predicts the future trajectories. Here $i$ represents the initialisation parameters for the motion model.}
    \label{fig:arch}
    \vspace{-0.35cm}
\end{figure*}

This section introduces the proposed hybrid model.
\mbox{Figure \ref{fig:arch}} shows the architecture of the hybrid model $g$, which contains the three main components: traffic scenario encoder $f_{\bm{\theta}}$, action decoder $d_{\bm{\vartheta}}$, and motion model $\varphi$.

\subsection{Problem Formulation}
\label{sec:problem_formulation}

The main objective in this work is to construct a hybrid \mbox{model $g$}, that is capable of predicting both the object trajectory $\mathcal{T}_{:T}^{m} $ and object attributes $\mathcal{A}_{:T}^{m} $  as time series over the time period $T$.
This is realised by the mapping
\begin{equation} 
  g: (\mathcal{O}_{:T_\text{hist}}^m, \bm{L}_{:T_\text{hist}},\textbf{\textit{I}}) \mapsto  \mathcal{O}_{:T_\text{pred}}^m,
\end{equation}
where \mbox{$\mathcal{O}_{:T}^{m} = (\mathcal{T}_{:T}^{m}, \mathcal{A}_{:T}^{m})$}. 
Here, $\mathcal{O}_{:T}^{m}$ represents the object state information of the investigated traffic participant $m$ within the traffic scenario for the respective time period $T$, which is either the past $T_{\text{hist}}$ or the future $T_{\text{pred}}$ to be predicted.
The environment state is represented by  \mbox{$ \bm{L}_{:T_{\text{hist}}} = { \left\{ \mathcal{O}_{:T_{\text{hist}}}^{m}\right\} }^{M}_{1} $}, which contains the object state information of all $M$ traffic participants within the scenario, and the infrastructure map information $\textbf{\textit{I}}$ in a \ac{bev} representation. $\bm{L}_{:T_\text{pred}}$ is formulated as follows,
\begin{equation}
    \bm{L}_{:T_\text{pred}} = \{\mathcal{O}_{:T_\text{pred}}^m = g(\mathcal{O}_{:T_\text{hist}}^m, \bm{L}_{:T_\text{hist}},\textbf{\textit{I}})~\forall~m  \in \left\{1,\dots,M\right\}\}.
\end{equation}

The hybrid model $g$ consists of the \ac{dl} models $f_{\bm{\theta}}$ and $d_{\bm{\vartheta}}$ along with the motion model $\varphi$. 
The \ac{dl} models are responsible for the mapping \mbox{$f_{\bm{\theta}}  \circ d_{\bm{\vartheta}}: (\mathcal{O}_{:T_\text{hist}}^m, \bm{L}_{:T_\text{hist}},\textbf{\textit{I}})\mapsto  \mathcal{A}_{:T_{\text{pred}}}^{m}$}, where ${\bm{\theta}}$ and $\bm{\vartheta}$ indicate the models with learnable parameters and \mbox{$ \mathcal{A}_{:T_{\text{pred}}}^{m} $} are the predicted action space outputs.
Within this action space the object behaviour is represented by features, \ie velocity, acceleration, and yaw rate.
The concrete features depend on the specific type of motion model $\varphi$ used. 
The motion model performs the mapping \mbox{$ \varphi: \mathcal{A}_{:T_{\text{pred}}}^{m} \mapsto \mathcal{T}_{:T_{\text{pred}}}^{m}$}, with the trajectory $\mathcal{T}_{:T_{\text{pred}}}^{m}$ containing the predicted positional information \mbox{$(x,y)$} of the $m^\text{th}$ traffic participant. 
The predicted action space features and the trajectories together represent the object state estimations \mbox{$\mathcal{O}_{:T_{\text{pred}}}^{m} = (\mathcal{T}_{:T_{\text{pred}}}^{m}, \mathcal{A}_{:T_{\text{pred}}}^{m})$} for $ T_{\text{pred}}$.

\subsection{Traffic Scenario Encoder}
\label{sec:scenario_encoder}
The traffic scenario encoder $f_{\bm{\theta}}$ is employed to encode the interacting traffic participant within a traffic scenario.
The interaction between traffic participants and their interactions with the infrastructure are crucial factors in predicting the behaviour of the traffic participants.
Consequently, both of these elements are incorporated into the traffic scenario encoder $f_{\bm{\theta}}$.
Furthermore, $f_{\bm{\theta}}$ is capable of processing a variable number of objects with differing sequence lengths.
The traffic participants of a scenario are represented by $\bm{L}_{:T_{\text{hist}}}$, which contains $M$ object state estimations  \mbox{$\mathcal{O}_{:T_{\text{hist}}}^{m} = (\mathcal{T}_{:T_{\text{hist}}}^{m}, \mathcal{A}_{:T_{\text{hist}}}^{m})$} including the object trajectory $\mathcal{T}_{:T_{\text{hist}}}^{m} $ and objects attributes  $\mathcal{A}_{:T_{\text{hist}}}^{m} $ in the form of  time series over the time period $T_{\text{hist}}$. 
Hereby, \mbox{$\mathcal{T}_{:T_{\text{hist}}}^{m} \in \mathbb{R}^{N_{\mathrm{p}} \times T_{\text{hist}}}$} with \mbox{$N_{\mathrm{p}} = 2 $} represents the positional information \mbox{$(x,y)$} and \mbox{$\mathcal{A}_{:T_{\text{hist}}}^{m} \in \mathbb{R}^{N_{\mathrm{a}} \times T_{\text{hist}}}$} with the number of object attribute features $N_{\mathrm{a}}$ \ie velocity, acceleration, and yaw rate.
The infrastructure information of each scenario is represented as \ac{bev} map image \mbox{$\textbf{\textit{I}} \in \mathbb{R}^{H\times W\times C}$}, with $H$, $W$, and $C$ being the height, width, and number of channels, respectively.
In total, each traffic scenario is represented by $M$ object state estimations ${ \left\{ \mathcal{O}_{:T_{\text{hist}}}^{m}\right\} }^{M}_{1}$ and the map \mbox{image $\textbf{\textit{I}}$}.
The main objective of the traffic scenario encoder is to embed and learn the relationship and interactions between the traffic participants and the map image $\textbf{\textit{I}}$.

The \ac{bev} map image $\textbf{\textit{I}}$ is processed by an infrastructure encoder $ f_{\textbf{\textit{I}}} $, realised by a ResNet-50~\cite{He2016}.
This model is used to perform the mapping \mbox{$f_{\textbf{\textit{I}}}: \textbf{\textit{I}} \mapsto \bm{z}^\text{I}$}.
In order to process all $M$ traffic participant within a scenario, $M$ encoder only transformer networks \cite{Vaswani2017} are applied to perform the mapping \mbox{$f_{\mathcal{O}}: \mathcal{O}_{:T_{\text{hist}}}^{m} \mapsto \bm{z}^{m}$} with each \mbox{$\bm{z} \in \mathbb{R}^{N_{\text{z}}}$}, \mbox{$N_{\text{z}} = 128$}.
The $M+1$ latent vectors $\bm{z}$ representing the m ap and the traffic participants are processed in parallel, with no interaction between the latent vectors. 
However, in order to develop a behaviour prediction model, it is necessary to model or learn the relationships between the various traffic participants as well as the relationships between the infrastructure and the aforementioned traffic participants.
In this work, a merger transformer model $f_{\text{merge}}$ with cross attention \cite{Vaswani2017} is employed to capture the relationships between the latent vectors of the traffic participants \mbox{$\left\{\bm{z}^\text{1},\ldots,\bm{z}^\text{M}\right\}$} and\mbox{map $\bm{z}^\text{I}$}.
Refining the individual embeddings by using $f_{\text{merge}}$  leads to contextually enriched representations \mbox{$\bm{H} \in  \mathbb{R}^{(M+1) \times N_{\text{h}}}$}.
As shown in Figure~\ref{fig:arch}, the $M+1$ latent vectors are passed through $f_\text{merge}$, where the latent vectors interact with each other through the attention mechanism.
Hereby, the mapping \mbox{$ f_{\text{merge}}: \left\{\bm{z}^\text{I}, \bm{z}^\text{1},\ldots,\bm{z}^\text{M}\right\} \mapsto \bm{H} = \left\{\bm{h}^\text{I}, \bm{h}^\text{1},\ldots,\bm{h}^\text{M}\right\} $} is performed where each \mbox{$\bm{h} \in \mathbb{R}^{N_{\text{h}}}$} has the same dimension \mbox{$N_{\text{h}} = 128$}.

Overall, the traffic scenario encoder creates the latent representation $ \bm{H}$ that incorporates the interactions between the traffic participants and the map, by \mbox{$f_{\bm{\theta}}: (\mathcal{O}_{:T_\text{hist}}^m, \bm{L}_{:T_\text{hist}},\textbf{\textit{I}}) \mapsto \bm{H}$}.

\subsection{Action Decoder}
\label{sec:action_decoder}
The goal of the action decoder $d_{\bm{\vartheta}}$ is to predict the actions space outputs $\bm{\mathcal{A}}_{:T_{\text{pred}}}$ from the latent embeddings \mbox{$\bm{H} = \left\{\bm{h}^\text{I}, \bm{h}^\text{1},\ldots,\bm{h}^\text{M}\right\}$}.
The action \mbox{decoder $d_{\bm{\vartheta}}$} is realised by a set of $K$ sequence decoders $d_k$, where $K$ is the absolute number of prediction steps.
All sequence decoders $d_k$ share the same weights and are implemented using the \ac{lstm} architecture \cite{Hochreiter1997}.
Each \ac{lstm} decoder performs the mapping \mbox{$d_k: 
(\textit{\textbf{q}}^{m}_{t}, \bm{L}_{t}) \mapsto  (\textit{\textbf{q}}^{m}_{t+1}, \mathcal{A}_{t+1}^{m})$} where \mbox{$\bm{q}^m_t \in \mathbb{R}^{N_{\text{h}}}$} is the hidden state of the \mbox{current timestamp ${t}$}.
This hidden state, along with the environment state $\bm{L}_{t}$, is used to predict the object attributes $\mathcal{A}_{t+1}^{m}$ for the next timestamp $t+1$. For the first \ac{lstm} decoder $d_1$, the current state $ \textit{\textbf{q}}^{m}_{t}$ is represented by the latent embeddings $\bm{H}$ and the environment state $\bm{L}_{t}$ by $\bm{L}_{:T_\text{hist}}$.

Overall, the action \mbox{decoder $d_{\bm{\vartheta}}$} performs the mapping \mbox{$d_{\bm{\vartheta}}: (\bm{H}, \bm{L}_{:T_\text{hist}}) \mapsto \bm{\mathcal{A}}_{:T_{\text{pred}}}$}, where \mbox{$\bm{\mathcal{A}}_{:T_{\text{pred}}}= \left\{\mathcal{A}^m_{:T_{\text{pred}}}\right\}_{m=1}^{M}$.}
It is a fundamental aspect of this concept to predict the object features $\mathcal{A}_{:T_{\text{pred}}}^{m}$ within the action space instead of directly predicting trajectories. 
Therefore, the plausibility of the predicted trajectories can be verified by the underlying actions space outputs.

Together, the traffic scenario encoder $f_{\bm{\theta}}$ and the action decoder $d_{\bm{\vartheta}}$ represent the learnable \ac{dl} components of the hybrid model architecture to perform the mapping 
\begin{equation}
    {f}_{\bm{\theta}} \circ {d}_{\bm{\vartheta}}: (\mathcal{O}_{:T_\text{hist}}^m, \bm{L}_{:T_\text{hist}},\textbf{\textit{I}}) \mapsto {\mathcal{A}}_{:T_{\text{pred}}}^{m}
\end{equation}
to predict the action \mbox{space outputs $\bm{\mathcal{A}}_{:T_{\text{pred}}}$}.

\subsection{Motion Model}
\label{sec:motion_model}
Similar to the action decoder $d_{\bm{\vartheta}}$, the kinematic motion model $\varphi$ is also constructed in a sequential design and consists of $K$ components $\varphi_k$. 
The motion model uses the output of the action \mbox{decoder $d_{\bm{\vartheta}}$}, the action space features, to predict the object trajectory \mbox{$\varphi_k: \bm{\mathcal{A}}_{:T_{\text{pred}}} \mapsto \bm{\mathcal{T}}_{:T_{\text{pred}}}$}, with \mbox{$\bm{\mathcal{T}}_{:T_{\text{pred}}} = \left[\mathcal{T}^m_{:T_{\text{pred}}}\right]_{m=1}^{M}$}.
The motion model $\varphi$ has no trainable parameters.
For this task, two distinct kinematic models are considered and presented in the following. 

\subsubsection{Constant Velocity Model}
For the \ac{cv} motion model, the \ac{dl} models ${f}_{\bm{\theta}} \circ {d}_{\bm{\vartheta}}$ predict the lateral and longitudinal velocities, \mbox{$\mathcal{A}_{:T_{\text{pred}}}^{m} = (\bm{v}_{\text{x}}^{m}, \bm{v}_{\text{y}}^{m})$}.
Hereby \mbox{$\bm{v}_{\text{x}}^{m} = \{ v_{\text{x}}^{m}(0), \dots, v_{\text{x}}^{m}(T_{\text{pred}})\}$} represents the longitudinal velocities of the $m^\text{th}$ traffic participant, and respectively $\bm{v}_{\text{y}}^{m}$ for the lateral velocities.
The \ac{cv} motion model uses $\bm{v}_{\text{x}}^{m}$ and $\bm{v}_{\text{y}}^{m}$ to predict the trajectory \mbox{$\mathcal{T}_{:T_{\text{pred}}}^{m} = (\bm{\hat{x}}^{m}, \bm{\hat{y}}^{m})$}.
It should be noted that the \ac{cv} motion model is executed every \mbox{timestep $t$} based on the current velocity values \mbox{$ \left(v_{\text{x}}^{m}(t), v_{\text{y}}^{m}(t)\right)$}, as the velocities are predicted for the entire prediction period $T_{\text{pred}}$.
The predicted trajectories and the velocity values are both presented in an egocentric coordinate system.

\subsubsection{Constant Turn Rate and Acceleration Model}
The \ac{ctra} motion model uses the yaw rate, also known as turn rate, and the acceleration, \mbox{$\mathcal{A}_{:T_{\text{pred}}}^{m} = (\bm{a}_{\text{x}}^{m},\bm{\dot{\psi}}^{m})$}.
Here, the acceleration is \mbox{$ \bm{a}^{m} = \{a^{m}(0), \dots, a^{m}(T_{\text{pred}}) \}$}, while the yaw rate is \mbox{$ \bm{\dot{\psi}}^{m} = \{\dot{\psi}^{m}(0), \dots, \dot{\psi}^{m}(T_{\text{pred}})\}$} for the $m^\text{th}$ traffic participants over $T_{\text{pred}}$.
The future trajectories \mbox{$\mathcal{T}_{:T_{\text{pred}}}^{m} = (\bm{\hat{x}}^{m}, \bm{\hat{y}}^{m})$} are predicted using the \ac{ctra} motion model.

Based on the given action space features, these kinematic models are capable to provide reliable predictions of the object trajectories for other traffic participants within the scenario. 
For further information on the \ac{cv} and \ac{ctra} model, interested readers may refer to \cite{Schubert2008} and \cite{Blackman1999}.

\subsection{Hybrid Model Training}
\label{sec:hybrid_model_training}
The proposed hybrid model is trained in an \mbox{end-to-end} fashion, with the loss $\mathcal{L}$ containing three learning objectives
\begin{equation}
    \mathcal{L} = \mathcal{L}_\text{MSE} +\mathcal{L}_\text{delta}+\mathcal{L}_\text{offroad}
\end{equation}
presented in the following. 

\subsubsection{MSE-Loss}
$\mathcal{L}_\text{MSE}$ is calculated by the \ac{mse} between the ground truth trajectory $(\bm{x}^{m}, \bm{y}^{m})$ and the predicted trajectory $(\bm{\hat{x}}^{m}, \bm{\hat{y}}^{m})$.

\subsubsection{Delta-Loss}
The second learning objective is referred to as delta-loss $\mathcal{L}_\text{delta}$ and constrains the action space output of the \ac{dl} models ${f}_{\bm{\theta}} $ and $ {d}_{\bm{\vartheta}}$.
$\mathcal{L}_\text{delta}$ constrains the predictions of $\mathcal{A}^m_{:T_\text{pred}}$ based on a minimal $\Delta^{\text{min}} \in \mathbb{R}^{N_{\mathrm{a}}}$ and maximal bound $\Delta^{\text{max}} \in \mathbb{R}^{N_{\mathrm{a}}}$, that are defined based on expert knowledge.
$\mathcal{L}_\text{delta}$ is defined as follows
\vspace{-0.15cm}
\begin{equation}
    \mathcal{L}_ {\text{delta}} = \sum_{m=1}^{M} \sum_{t=1}^{T_{\text{pred}}}  {\delta}^m_t,
    \label{Eq:TPII_delta}
\end{equation}
\begin{equation}  
        {\delta}^m_t = 
        \begin{cases}
            0  & \text{if } \Delta^{\text{min}}_{i} \leq \mathcal{A}^{m}_{t}[i] \leq \Delta^{\text{max}}_{i}~\forall~i \in \{1, \dots, N_a \}\\
            1  & \text{otherwise}
        \end{cases}.
\end{equation}
The bounds $[\Delta^{\text{min}}; \Delta^{\text{max}}]$ are defined using expert knowledge to satisfy the physical constraints of vehicle dynamics \cite{ Botsch2020, Yusof2016}.
For instance, the absolute acceleration for vehicles does exceed \SI{8}{\meter\per\second\squared} in normal driving situations \cite{gillespie1992, Ristic2006}.
This loss component applied to the action space is an essential component of the presented method. 
The loss shapes the action space of the objects in the traffic scenario, and not the final trajectory space.

\subsubsection{Offroad-Loss}
The third learning objective used in this work is referred to as offroad loss $\mathcal{L}_\text{offroad}$. 
It is used to penalise the network for generating trajectories outside the drivable area. 
To realise this, the predicted trajectories $\bm{\mathcal{T}}_{:T_{\text{pred}}}$ are converted into a binary image $\bm{I}_\mathrm{traj} \in \mathbb{R}^{H \times W}$, with $H$ and $W$ being the heigth and width of $\bm{I}_\mathrm{traj}$.
Additionally, the infrastructure map $\bm{I}$ is used to create a second binary \mbox{image $\bm{I}_{\text{driveable}} \in \mathbb{R}^{H \times W}$}, where the drivable region is represented by the value $1$ and the offroad region by $0$. 
$\mathcal{L}_\text{offroad}$ uses the difference between the two binary maps
\vspace{-0.2cm}
\begin{equation}
    \mathcal{L}_\text{offroad} = \sum_{i=1}^{H} \sum_{j=1}^{W} \max \left(0, \bm{I}_\text{traj}(i,j) - \bm{I}_{\text{driveable}}(i,j)\right),
\end{equation}
 to identify trajectory elements, which are outside the drivable region.

In summary, this work introduces a novel hybrid model, combining modern \ac{dl} models with a traditional motion model.
Hereby, the applied losses enforce the network to consider the physical constraints of vehicle dynamics in the action space and predict realistic trajectories of traffic participants.
An important contribution of this work is the application of physical constraints on the action space with specific expert knowledge. 
This is an effective method in order to increase the trustworthiness of \ac{dl} methods and promote safe path-planning procedures.

\newcommand{\getSubFigSizeFigTwo}{0.19}
\setcounter{subfigure}{0}
\begin{figure*} [th!]
    \centering
    \setcounter{subfigure}{0}
    \renewcommand\thesubfigure{(\alph{subfigure})}
    \subfigure[][]{
        \includegraphics[width=\getSubFigSizeFigTwo\textwidth]{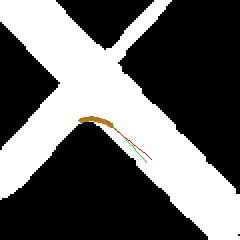}
    }
    \subfigure[][]{
        \includegraphics[width=\getSubFigSizeFigTwo\textwidth]{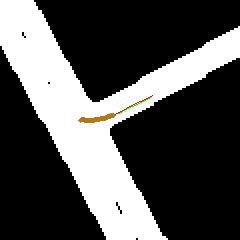}
    }
    \subfigure[][]{
        \includegraphics[width=\getSubFigSizeFigTwo\textwidth]{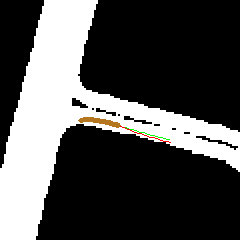}
    }
    \subfigure[][]{
        \includegraphics[width=\getSubFigSizeFigTwo\textwidth]{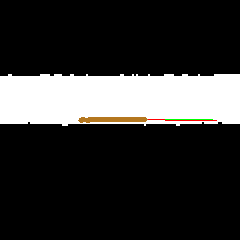}
    }
    \subfigure[][]{
        \includegraphics[width=\getSubFigSizeFigTwo\textwidth]{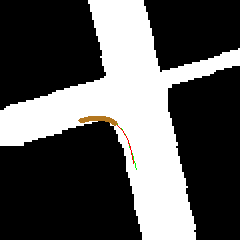}
    }
    \subfigure[][]{
        \includegraphics[width=\getSubFigSizeFigTwo\textwidth]{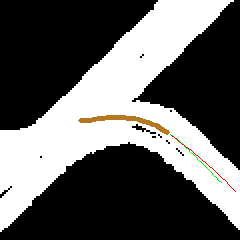}
    }
    \subfigure[][]{
        \includegraphics[width=\getSubFigSizeFigTwo\textwidth]{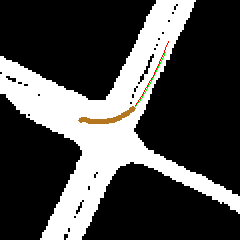}
    }
    \subfigure[][]{
        \includegraphics[width=\getSubFigSizeFigTwo\textwidth]{858_6.png}
    }

    \caption{
            Eight driving scenarios from the Argoverse validation set are shown.
            The history information is available for \mbox{$T_{\text{hist}} = \SI{2}{\second}$} and coloured in orange.
            The model's prediction of the EGO vehicle's trajectory is made for \mbox{$T_{\text{pred}} = \SI{3}{\second}$} and shown in red, while the corresponding ground truth trajectories are shown in green.
        }
        \label{fig:tprediction}
     \vspace{-0.5cm}
\end{figure*}


\section{Experiments and Results} \label{Sec:experiments}
In this section, the experiments and results of the proposed method are discussed. 
With the experiments, the following questions are addressed:
\begin{enumerate}
     \item  \textit{How accurate are the trajectories predicted by the hybrid model?}
     \item  \textit{Is a hybrid model beneficial for trajectory prediction?}
     \item  \textit{Is the generated action space plausible?}    
     \item  \textit{How can safe planning procedures benefit from the proposed architecture?}
 \end{enumerate}

\subsection{Dataset}
To train and evaluate the proposed hybrid model, the established and publicly available \mbox{Argoverse 1} motion forecasting dataset \cite{Chang2019} for trajectory prediction is used.
This dataset contains real-world traffic data recorded from an EGO vehicle.
The train and validation splits contain 205,942 and 39,472 traffic scenarios, respectively.
Each scenario is recorded for \SI{5}{\second} at a frequency of \SI{10}{\hertz}.
For the task of trajectory prediction, the trajectories and action space features are given for $T_{\text{hist}} = \SI{2}{\second}$ and predicted for \mbox{$T_\text{pred} = \SI{3}{\second}$}.
The results reported are based on experiments on the Argoverse validation dataset.

\subsection{Implementation Details} \label{sec:experiments:implementation_details}
The hybrid model\footnote{An implementation of the proposed method is available at: \mbox{\url{https://github.com/MB-Team-THI/hmcas}}} is trained using two distinct training configurations.
For both configurations the following training settings are used: batch size $b=64$, number of \mbox{epoch $n_{epochs} = 150$}, learning rate $\alpha=0.001$ and Adam optimiser \cite{Kingma2015}. 

\subsubsection{\ac{hss}} 
In this training configuration, as illustrated in Figure~\ref{fig:arch}, the model makes a \textit{single} prediction for the whole time prediction period \mbox{$T_\text{pred} = \SI{3}{\second}$}. 
Hereby, the prediction includes the action space and trajectories of all $M$ traffic participants within the scenario. 
This prediction is based on the data of length $T_{\text{hist}} = \SI{2}{\second}$ representing the traffic scenario as input, resulting in the following mapping
\begin{equation}
      g:  \mathcal (\mathcal{O}_{:\SI{-2}{\second}}^m, \bm{L}_{:\SI{-2}{\second}},\textbf{\textit{I}}) \mapsto {O}_{:\SI{3}{\second}}^{m}.
\end{equation}

\subsubsection{\ac{hms}}
In contrast to the \ac{hss} configuration, in \ac{hms} the prediction is divided into \textit{multiple} steps. 
The prediction period here is $T_\text{pred} = \SI{1}{\second}$, which corresponds to 10 timesteps at the given frequency of \SI{10}{\hertz}, while the input length is $T_\text{hist} = \SI{2}{\second}$, summarised as
\begin{equation}
      g:  \mathcal (\mathcal{O}_{:\SI{-2}{\second}}^m, \bm{L}_{:\SI{-2}{\second}},\textbf{\textit{I}}) \mapsto {O}_{:\SI{1}{\second}}^{m}.
\end{equation}
In the subsequent prediction step, the prediction results are concatenated with the previous input, omitting the first second of the history information $T_\text{hist}$ to maintain a constant input size of \SI{2}{\second}.
Accordingly, in the second step of the \ac{hms} configuration, the history information is updated with the preceding  prediction results.
This process is repeated for the prediction cycle until the trajectories for \SI{3}{\second} are available.
The underlying assumption of this configuration is that the model utilises the provided prediction as an input, thereby enabling it to rectify any errors in the initial prediction.

\subsection{Baselines and Metrics}
To evaluate the proposed method, a benchmark comparison is performed that incorporates different types of models. 
This includes \ac{dl} models like VectorNet~\cite{Gao2020}, HOME~\cite{Gilles2021} and the \ac{lstm} implementation from \cite{Chang2019}.
Several variations of the proposed architecture are examined through an ablation study, using \ac{ade} and \ac{fde} to evaluate predicted trajectories \cite{Rudenko2020}, and assessing physical feasibility by analysing the acceleration of the predicted object features.
\begin{table}[b]
    \vspace{-0.3cm}
    \caption{Benchmark comparison by \ac{ade} and \ac{fde} on the Argoverse validation dataset.}
    \centering   
    \setlength\extrarowheight{1pt}
    \begin{tabular}{|c|c|c|c|}
        \hline
        Model-Family                    & Method                    & \ac{ade} [\SI{}{\meter}]&  \ac{fde} [\SI{}{\meter}]\\
        \hline
        \hline
        Baseline-Model                  & \ac{cv}                   & 3.53 & 7.78 \\
        \hline
        \multirow{4}{*}{\ac{dl}-Model}  & \ac{lstm} \cite{Chang2019}                & 2.15  & 4.95 \\
                                        & VectorNet \cite{Gao2020}  & 1.81  & 4.01 \\
                                        & HOME \cite{Gilles2021}    & -     & 3.81 \\ 
                                        & \acs{wmm}                 & 2.2   & 5.75 \\
        \hline
        \multirow{4}{*}{Hybrid-Model}   & \ac{hss}+\ac{cv}          & 1.73  & 4.00 \\
                                        & \ac{hss}+\ac{ctra}        & 1.67  & 3.71 \\
                                        & \ac{hms}+\ac{cv}          & 1.66  & 3.68 \\
                                        & \ac{hms}+\ac{ctra}        & \textbf{1.60} & \textbf{3.65} \\
        \hline
    \end{tabular}
    \label{tab:TPII_displacement_error}
\end{table}

\subsection{Results}
\subsubsection{How accurate are the behaviours predicted by the hybrid model?}
To answer this question, the trajectories of the EGO vehicle are evaluated over the \mbox{prediction time $T_{:\text{pred}}$.}
The predicted trajectories of the EGO vehicle $\mathcal{\hat{T}}^{\text{ego}}_{T_{:\text{pred}}}$ are compared against the original ground truth trajectories of the EGO vehicle $\mathcal{T}^{\text{ego}}_{T_{:\text{pred}}}$.
The ground truth trajectories represent actual human driving behaviour and can therefore be used to assess how realistic the predicted trajectories are, as shown in Figure \ref{fig:tprediction} for eight scenarios.
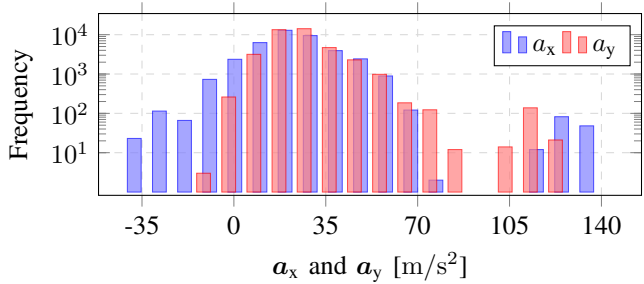
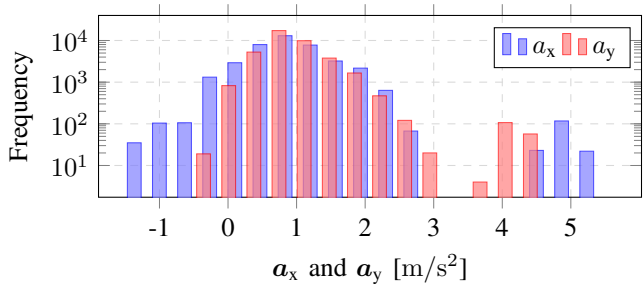
\begin{figure}[b!]
    \vspace{-0.5cm}
    \centering
    \subfigure[Histogram of  $\bm{a}_{\text{x}}$ and $\bm{a}_{\text{y}}$ without $\mathcal{L}_\text{delta}$]{
        \begin{tikzpicture}
            \begin{axis}[
                ybar,
                ymode=log, 
                width=8.8cm, height=4cm,
                xlabel={$\bm{a}_{\text{x}}$ and $\bm{a}_{\text{y}}$ [\SI{}{\meter\per\second\squared}]},
                ylabel={Frequency},
                legend style={at={(0.98, 0.9354)}, legend columns=-1},
                grid=major,
                grid style={dashed, gray!30},
                bar width=5.2pt,
                enlarge x limits=0.1,
                ymin=0,
                ytick={10, 100, 1000, 10000},
                yticklabels={$10^1$, $10^2$, $10^3$, $10^4$}, 
                xtick={-35, 0, 35, 70, 105, 140}, 
                xticklabels={-35, 0, 35, 70, 105, 140}, 
                ]            
                \addplot[blue, fill=blue!50, opacity=0.7] 
                table[x=Value, y=Frequency, col sep=comma] {without_l_delta_ax_histogram.csv};
                \addlegendentry{$a_{\text{x}}$}
                \addplot[red, fill=red!50, opacity=0.7]
                table[x=Value, y=Frequency, col sep=comma] {without_l_delta_ay_histogram.csv};
                \addlegendentry{$a_{\text{y}}$}
            \end{axis}
        \end{tikzpicture}     
    }
    
    \subfigure[Histogram of  $\bm{a}_{\text{x}}$ and $\bm{a}_{\text{y}}$ with $\mathcal{L}_\text{delta}$]{        
        \begin{tikzpicture}
            \begin{axis}[
                ybar,
                ymode=log, 
                width=8.8cm, height=4cm,
                xlabel={$\bm{a}_{\text{x}}$ and $\bm{a}_{\text{y}}$ [\SI{}{\meter\per\second\squared}]},
                ylabel={Frequency},
                legend style={at={(0.98, 0.9354)}, legend columns=-1},
                grid=major,
                grid style={dashed, gray!30},
                bar width=5.2pt,
                enlarge x limits=0.1,
                ymin=0,
                ytick={10, 100, 1000, 10000},
                yticklabels={$10^1$, $10^2$, $10^3$, $10^4$}, 
                xtick={-1, 0, 1, 2, 3, 4, 5}, 
                xticklabels={-1, 0, 1, 2, 3, 4, 5} 
                ]
                \addplot[blue, fill=blue!50, opacity=0.7] 
                table[x=Value, y=Frequency, col sep=comma] {with_l_delta_ax_histogram.csv};
                \addlegendentry{$a_{\text{x}}$}    
                \addplot[red, fill=red!50, opacity=0.7] 
                table[x=Value, y=Frequency, col sep=comma] {with_l_delta_ay_histogram.csv};
                \addlegendentry{$a_{\text{y}}$}
            \end{axis}
        \end{tikzpicture}
        
    }
    \caption{
    The histograms illustrate the maximum predicted $\bm{a}_{\text{x}}$ and $\bm{a}_{\text{y}}$ of the EGO vehicle per scenario for the Argoverse validation dataset with and without the delta-loss $\mathcal{L}_\text{delta}$, thereby demonstrating the constraining capabilities of $\mathcal{L}_\text{delta}$ on the action space.
    Using $\mathcal{L}_\text{delta}$ the range of $\bm{a}_{\text{x}}$ and $\bm{a}_{\text{y}}$ was reduced from $\SIrange{-34}{138}{\meter\per\second\squared}$ to $\SIrange{-1.23}{5.37}{\meter\per\second\squared}$. 
    }
    \label{fig:a_histogram}
\end{figure}

The proposed hybrid model is trained in two training configurations, \ac{hss} and \ac{hms}, as described in \mbox{Section \ref{sec:experiments:implementation_details}}.  
Furthermore, two variants of the motion model are considered, the \ac{cv} and \ac{ctra} model.
Additionally, the proposed hybrid model is once trained without the motion model $\varphi$, whereby the trajectories are directly predicted by the \ac{dl} \mbox{models $f_{\bm{\theta}}  \circ d_{\bm{\vartheta}}$}, in order to gain insights into the effect of $\varphi$, which is termed as \ac{wmm}. 
For comparison, also the \ac{cv} model is used for trajectory prediction based on the previous states of the EGO vehicle \mbox{(cf. Section \ref{sec:motion_model})}.
The investigated combinations and models are named as follows:
\begin{itemize}
    \item \textbf{\ac{hss}+\ac{cv}:} \acl{hss} with \ac{cv} model,
    \item \textbf{\ac{hss}+\ac{ctra}:} \acl{hss} with \ac{ctra} model,
    \item \textbf{\ac{hms}+\ac{cv}:} \acl{hms} with \ac{cv} model,
    \item \textbf{\ac{hms}+\ac{ctra}:} \acl{hms} with \ac{ctra} model,
    \item \textbf{\ac{cv}:} Constant velocity model, and
    \item \textbf{\ac{wmm}:} Hybrid model without the motion model $\varphi$.
    
\end{itemize}

\newcommand{\getSubFigSizeFigThree}{0.17}
\begin{figure}[b!]
    \vspace{-0.5cm}
    \centering
    \setcounter{subfigure}{0}
    \renewcommand\thesubfigure{(\alph{subfigure})}
    \subfigure[][]{\label{fig:fig3:sufig1}
        \setcounter{subfigure}{0}
        \renewcommand\thesubfigure{(\roman{subfigure})}
        \subfigure[Trajectory]{
            \includegraphics[width=\getSubFigSizeFigThree\textwidth]{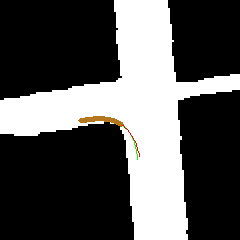}          
        }      
        \hfill
        \subfigure[$(\bm{a}_{\text{x}}, \bm{a}_{\text{y}})$]{
            \includegraphics[width=\getSubFigSizeFigThree\textwidth]{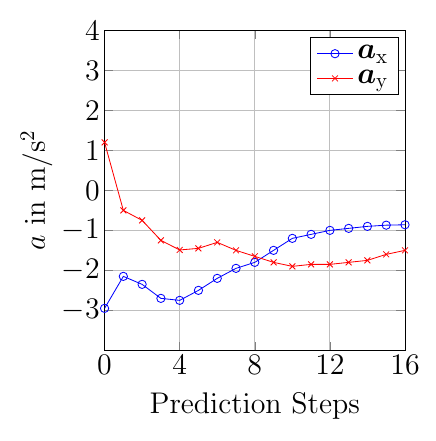}          
        }   
        \addtocounter{subfigure}{-1}
    } 
    \setcounter{subfigure}{1}
    \renewcommand\thesubfigure{(\alph{subfigure})}   
    \subfigure[][]{\label{fig:fig3:sufig2} 
        \setcounter{subfigure}{0}
        \renewcommand\thesubfigure{(\roman{subfigure})}
        \subfigure[Trajectory]{
            \includegraphics[width=\getSubFigSizeFigThree\textwidth]{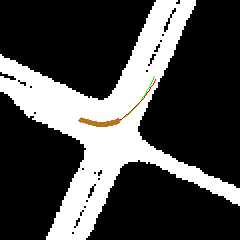}          
        }      
        \hfill
        \subfigure[$(\bm{a}_{\text{x}}, \bm{a}_{\text{y}})$]{
            \includegraphics[width=\getSubFigSizeFigThree\textwidth]{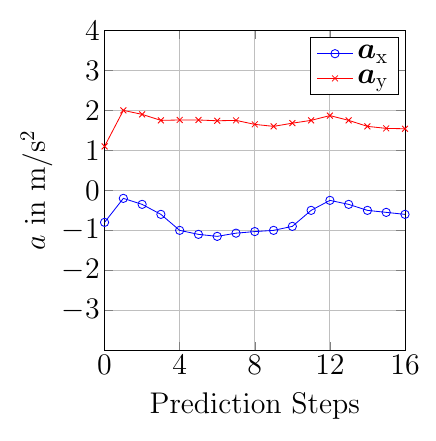}          
        }
    }  
    \caption{In (i) driving scenarios from the Argoverse validation set are shown. The history information is available for \mbox{$T_{\text{hist}} = \SI{2}{\second}$} and coloured in orange. The model's prediction of the EGO vehicle's trajectory is made for \mbox{$T_{\text{pred}} = \SI{3}{\second}$} and shown in red, while the corresponding ground truth trajectories are shown in green. In (ii) the longitudinal acceleration $\bm{a}_{\text{x}}$ and lateral acceleration $\bm{a}_{\text{y}}$ are shown.}
    \vspace{-0.1cm}
    \label{fig:TPII_actionspace}
\end{figure}

The ablation study resulting from these different training configurations is presented in \mbox{Table~\ref{tab:TPII_displacement_error}}.
Additionally, also other trajectory prediction models are included to enable a baseline comparison. 

The \ac{hms}+\ac{ctra} model configuration performs best compared to the other variations of the hybrid model-based on the metrics \ac{ade} and \ac{fde}. Thus, the hybrid model effectively captures realistic and accurate driving behaviour, thereby providing a positive response to the first research question.

\subsubsection{Is a hybrid model beneficial for trajectory prediction?}
The proposed hybrid model combines deep learning models ($f_{\bm{\theta}} \circ d_{\bm{\vartheta}}$) with the motion model $\varphi$ for trajectory prediction. The ablation study within Table~\ref{tab:TPII_displacement_error} highlights the advantages of this approach. The \ac{cv} model alone is not competitive, and the \ac{wmm} suffers from a significant performance drop, highlighting the importance of $\varphi$ in the hybrid model. Moreover, the integration of the kinematic motion model improves the performance of deep learning models. In conclusion, the hybrid model improves both interpretability and performance in trajectory prediction tasks.

\subsubsection{Is the generated action space plausible?}
A key feature of the proposed hybrid model is the use of multiple output spaces, particularly the action output space $\mathcal{A}_{T_{\text{pred}}}$ \mbox{(cf. Sec. \ref{sec:motion_model})}. 
To evaluate the impact of the included expert knowledge within the delta loss $\mathcal{L}_\text{delta}$ (Eq.~\ref{Eq:TPII_delta}), models are trained with and without it. 
Figure~\ref{fig:a_histogram} shows that the model trained with $\mathcal{L}_\text{delta}$ produces plausible vehicle acceleration values, while the model trained without $\mathcal{L}_\text{delta}$ predicts physically infeasible values.
This highlights the value of expert knowledge in generating a realistic action space.
Upon examination of the histograms in Figure~\ref{fig:a_histogram}, it becomes notable that the impact of $\mathcal{L}_{\text{delta}}$ bears resemblance to that of scaling, yet is nevertheless distinct.
Figure~\ref{fig:TPII_actionspace} further illustrates that the predicted acceleration values are smooth and result in realistic trajectories.

\subsubsection{\textit{How can safe planing procedures benefit from the proposed architecture?}}
The proposed hybrid model offers two key advantages that are particularly relevant for the development of safe planning procedures.
Firstly, the utilisation of a well-established and robust kinematic motion model within the architecture facilitates the creation of reliable drivable trajectories based on intermediate prediction within the action space.
Secondly, to further ensure the reliability of the action space predictions, expert knowledge is incorporated within the learning objective to create an action space based on the physical constraints of vehicle dynamics.
This therefore improves the interpretability of the action space and thus the trustworthiness of such prediction methods.
In summary, safe planning procedures can benefit from these two key components of the proposed hybrid model.

\section{Conclusion} \label{Sec:conclusion}
This work introduces a novel hybrid model for trajectory prediction. 
The proposed architecture combines modern \ac{dl} models and established motion models in an end-to-end trainable manner.
By the incorporation of expert knowledge into the learning objective, the physical constraints of vehicle dynamics are considered within the predicted object attributes and trajectories. 
Therefore, this hybrid model is capable of shaping a realistic action space and predicting trajectories that are reliably drivable.
These characteristics are fundamental for establishing the desired high level of trust in autonomous vehicles and for promoting safer planning procedures.
Experiments on the publicly available real-world Argoverse dataset show that the proposed hybrid model is capable to generate realistic driving behaviour and outperforms other \ac{dl} approaches.
In addition, the positive effects of the various key components have been quantitatively and qualitatively demonstrated in ablation studies. 
This paper demonstrates how the incorporation of expert knowledge into \ac{dl} models enables the creation of a physically realistic action space and thus the robust prediction of trajectories.

\section*{Acknowledgement} 
\label{Sec:acknowledgement}
This work is supported by the Federal Ministry of Education and Research of Germany (BMBF) in the framework of FH-Impuls (project number 13FH7I13IA).

\bibliographystyle{IEEEtran}
\bibliography{arxiv_lib_file}

\end{document}